\documentclass[sigconf]{acmart}

\usepackage{times}
\usepackage{latexsym}
\usepackage{multirow}
\usepackage{booktabs}

\usepackage{microtype}
\usepackage{graphicx}
\usepackage{subcaption}

\AtBeginDocument{%
  \providecommand\BibTeX{{%
    \normalfont B\kern-0.5em{\scshape i\kern-0.25em b}\kern-0.8em\TeX}}}

\setcopyright{acmcopyright}
\copyrightyear{2021}
\acmYear{2021}
\acmDOI{10.1145/1122445.1122456}

\acmConference[CIKM '21]{CIKM '21: 30th ACM International Conference on Information and Knowledge Management}{November 01--05, 2021}{Queensland, Australia}
\acmBooktitle{CIKM '21: 30th ACM International Conference on Information and Knowledge Management, November 01--05, 2021, Queensland, Australia}
\acmPrice{15.00}
\acmISBN{978-1-4503-XXXX-X/18/06}

\begin{document}

\title{VT-SSum: A Benchmark Dataset for Video Transcript Segmentation and Summarization}

\author{Tengchao Lv}
\email{tengchaolv@pku.edu.cn}
\authornote{Work done during internship at Microsoft Research Asia.}
\affiliation{
  \institution{Peking University}
  \city{Beijing}
  \country{China}
}

\author{Lei Cui}
\email{lecu@microsoft.com}
\affiliation{
  \institution{Microsoft Research Asia}
  \city{Beijing}
  \country{China}
}

\author{Momcilo Vasilijevic}
\email{movasi@microsoft.com}
\affiliation{
  \institution{Microsoft}
  \city{Belgrade}
  \country{Serbia}
}

\author{Furu Wei}
\email{fuwei@microsoft.com}
\affiliation{
  \institution{Microsoft Research Asia}
  \city{Beijing}
  \country{China}
}

\renewcommand{\shortauthors}{Anonymous Submission}

\begin{abstract}
 Video transcript summarization is a fundamental task for video understanding. Conventional approaches for transcript summarization are usually built upon the summarization data for written language such as news articles, while the domain discrepancy may degrade the model performance on spoken text. In this paper, we present \textbf{VT-SSum}, a benchmark dataset with spoken language for video transcript segmentation and summarization, which includes 125K transcript-summary pairs from 9,616 videos. VT-SSum takes advantage of the videos from VideoLectures.NET by leveraging the slides content as the weak supervision to generate the extractive summary for video transcripts. Experiments with a state-of-the-art deep learning approach show that the model trained with VT-SSum brings a significant improvement on the AMI spoken text summarization benchmark. VT-SSum is publicly available at \url{https://github.com/Dod-o/VT-SSum} to support the future research of video transcript segmentation and summarization tasks.
\end{abstract}




\maketitle

\section{Introduction}

Video understanding has recently been a trending research topic because it involves a set of cross-modality problems and is an ideal testbed for CV and NLP models. The transcript summarization is a fundamental task for video understanding due to its wide application in video chaptering and keyframes/shots extraction by making unique contributions from the language/textual perspective. Existing text summarization approaches are usually built upon the deep learning models that are trained with datasets for written language such as news articles. There are two problems when applying current approaches to spoken language. First, the domain discrepancy between written language and spoken language may degrade the model performance~\cite{savelieva2020abstractive}. Second, video transcripts are often much longer than news articles, thereby the models trained with news articles may have never seen the lengthy input from transcripts. These make the current text summarization models overstretched on the spoken language tasks. Therefore, it is inevitable to create an in-domain summarization dataset for the spoken language, especially the video/audio transcripts.
In recent years, text summarization models are usually trained with the popular CNN/DailyMail~\cite{hermann2015teaching,nallapati2016abstractive}, Gigaword ~\cite{rush-etal-2015-neural}, XSum~\cite{narayan2018dont}, as well as the DUC 2004 dataset. These publicly available summarization datasets usually take advantage of news highlights or news headlines as the weak supervision, thus alleviating the human labeling efforts that are labor-intensive and time-consuming. Although these datasets help push the SOTA results significantly, all these are built for written texts while leaving spoken language summarization data unexplored for a long time. Therefore, to facilitate the research for spoken text summarization with minimum labeling efforts, it is indispensable to leverage the weak supervision to create a high quality dataset with sufficient training samples.

\begin{figure*}[ht]
    \centering
    \includegraphics[width=1\textwidth]{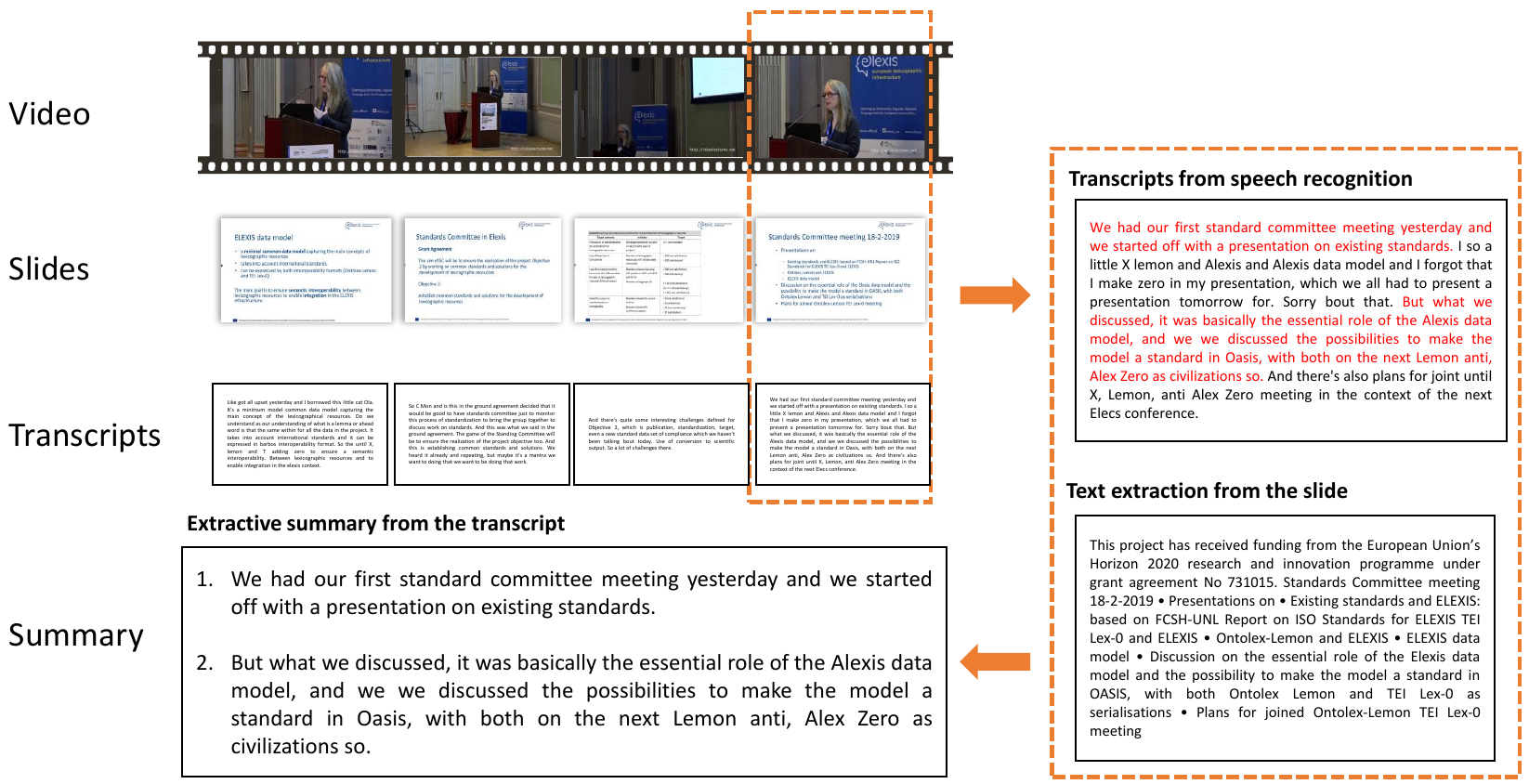}
    \caption{An illustration of using VideoLectures.NET to create page-level extractive summarization for video transcripts, where the presentation videos are first segmented into a list of coherent segments according to the slides switches, and transcript sentences and slides content are accurately aligned to generate the VT-SSum dataset using the extractive training method.}
    \label{fig:1}
\end{figure*}

To this end, we present VT-SSum, a benchmark dataset for video transcript segmentation and summarization. Inspired by the existing work, we leverage a simple but effective approach to obtain high quality spoken text summarization data with weak supervision. Specifically, we use the video transcripts and its corresponding slides from VideoLectures.NET to generate the extractive summary, which is shown in Figure~\ref{fig:1}. VideoLectures.NET is the world's biggest academic online video repository mainly targeted for the computer science area, but also includes videos in other domains such as social science, business, medicine, biology and many others. In VideoLectures.NET, each presentation usually includes the video as well as its corresponding slides. Meanwhile, the timeline of video and slides are accurately aligned so that the video/audio can be strictly segmented for each slides page. To generate the extractive summary, we assume that the text content of each slides page reflects the main idea of the presentation utterance, thus the extractive training method~\cite{10.5555/3298483.3298681} can be used to leverage the slides content as the weak supervision. In this way, we have obtained 9,616 videos, which further produces 125K transcript-summary pairs. To verify the effectiveness of VT-SSum, we conduct experiments with a deep learning model to compare the model performance of written and spoken text summarization. Experiment results confirms the domain mismatch between the written and spoken text summarization models, and show that the model trained with VT-SSum significantly outperforms the previous models on two testing datasets for spoken text summarization.

The contributions of the paper are summarized as follows:
\begin{itemize}
    \item We present VT-SSum, a benchmark dataset for video transcript segmentation and summarization that includes 125K transcript-summary pairs. To the best of our knowledge, this is the first summarization dataset for spoken language constructed with weak supervision.
    \item we conduct extensive experiments with deep learning models to confirm the effectiveness of VT-SSum for spoken text summarization.
    \item The VT-SSum dataset is publicly available at \url{https://github.com/Dod-o/VT-SSum}.
\end{itemize}

\section{VT-SSum}

In this section, we introduce the data creation pipeline for VT-SSum. Basically, the pipeline includes four steps: 1) Data collection; 2) Audio separation from the video and speech recognition; 3) Text extraction from slides; 4) Transcript segmentation; 5) Aligning the transcript and slides for extractive training. 

\subsection{Data Collection}

We use the Selenium toolkit\footnote{\url{https://www.selenium.dev/}} to crawl the videos from VideoLectures.NET, where 9,616 videos from 26 categories are totally obtained. For each presentation page, it usually contains the video content, presentation slides as well as a timeline to align the video and slides. For the sake of data integrity,  we only keep the presentation pages if and only if it contains both the video and slides, while filtering other presentation pages. As the format of slides includes ``.ppt'', ``.pptx'' and ``.pdf'', we convert all of them to the ``.pdf'' format that is easier to extract the text information.   

\subsection{Audio Separation and Speech Recognition}

We use the MoviePy toolkit\footnote{\url{https://zulko.github.io/moviepy/}} to separate the audio information from the videos. As the timeline for each presentation strictly denotes the alignment of video/audio and slides, we segment the whole audio into a sequence of audio clips according to the timestamp of the slides switch, so that each audio clip is accurately aligned to a slides page. To generate the transcripts for each slides page, we use Microsoft Speech to Text\footnote{\url{https://azure.microsoft.com/en-us/services/cognitive-services/speech-to-text/}} to convert each audio clip into a set of sentences.

\subsection{Text Extraction from Slides}

In VideoLectures.NET, the slides for each presentation are converted into a sequence of images that are aligned to the videos. Due to the low resolution of the images, simply extracting text information using OCR may bring incorrect tokens and is not acceptable. Therefore, we need to align the presentation slides in PDF with the converted images so that we can further extract text information from the digital-born PDF files. However, presenters may not start from the first page and may also skip some pages during the presentation, which requires additional efforts to align the PDF and slides images. To this end, we adopt the ORB~\cite{6126544} algorithm to measure the visual similarity between each PDF page and each slides image from the website, so that they can be accurately aligned for text extraction. Finally, we use the PyMuPDF toolkit\footnote{\url{https://github.com/pymupdf/PyMuPDF}} to extract the texts from the PDF file.

\subsection{Transcript Segmentation}

Text segmentation aims to divide a text block into contiguous segments based on its topics and semantic structures, which is a longstanding challenge in language understanding. For video transcript summarization, we assume that video transcripts can be organized into a list of coherent segments, where salient sentences can be extracted from each segment to compose the summary for the whole video. In this work, it is naturally believed that the transcript content for each presentation slide from the video is highly coherent, meanwhile the main idea is conveyed through the text information in the corresponding slide page. Therefore, we create transcript segmentation annotations by grouping transcript sentences within each presentation slide, where the timeline for slide switch and the video transcript are strictly aligned. In this way, the whole video transcript is annotated with the segmentation boundaries, which further facilitates the extractive training with the slides-transcripts alignment information.

\subsection{Aligning the Transcript and Slides for Extractive Training}

Given the transcript for each video clip and extracted texts from the corresponding slides page, we use the extractive training method~\cite{10.5555/3298483.3298681} to obtain the transcript-summary pairs. Distinct from the existing summarization dataset from news articles, we do not have the ground-truth summary from the news editors. Therefore, we assume that the main idea of the presentation is expressed in the slides content, thereby it can be leveraged as the weak supervision the produce the extractive summary. Specifically, we consider the slides content as the gold summary for the video clip, where the selected sentences from the transcript are the ones that maximize the ROUGE score with respect to the gold summary. To obtain the high quality transcript-summary pairs, we remove the slides with only pictures or very few words (less than 10 words) where the gold summary is not reliable. Meanwhile, we also set a threshold for the ROUGE score in order to filter the samples where transcripts and slides are not correctly matched. 

\subsection{Statistics and Baseline}

\begin{figure}[ht]
\centering
    \begin{subfigure}[b]{0.23\textwidth}
        \includegraphics[width=\textwidth]{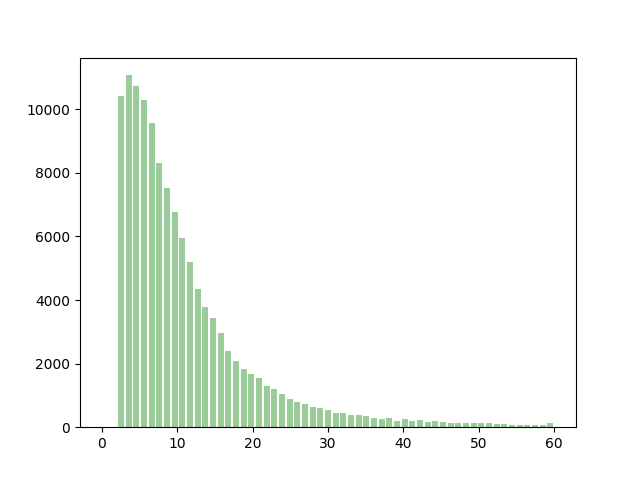}
        \caption{}
        \label{fig:2a}
    \end{subfigure}
    \begin{subfigure}[b]{0.23\textwidth}
        \includegraphics[width=\textwidth]{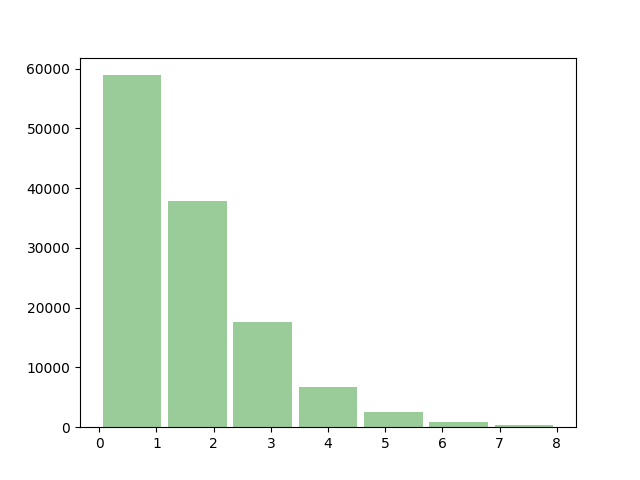}
        \caption{}
        \label{fig:2b}
    \end{subfigure}
    \caption{The length distribution of the input text (a) and summary (b), where the x-axis denotes the number of sentences and y-axis denotes the number of samples}\label{fig:2}
\end{figure}

We have totally collected 9,616 videos which produce 125K transcript-summary pairs. The statistics are shown in Table~\ref{tab:1}. Although the length of whole video transcript is much longer, the average length for VT-SSum is much smaller than the CNN/DM dataset due to the segmentation by slides page as shown in Table~\ref{tab:2}, and the length distribution of the text and summary are shown in Figure~\ref{fig:2}. Therefore, VT-SSum can fit into the current Transformer-based models with the length limitation. In particular, we use the PreSumm model~\cite{liu-lapata-2019-text} as the baseline approach to evaluate the performance of VT-SSum. PreSumm is a simple but effective summarization model based on the Transformer architecture. The extractive model is built on top of BERT~\cite{devlin-etal-2019-bert} by stacking several inter-sentence Transformer layers, which achieves the SOTA results on several benchmark datasets.

\begin{table}[t]
    \centering
    \begin{tabular}{cccc}
    \toprule
        \bf{Dataset} &  \bf{Avg. slides} & \bf{Avg. sents} & \bf{Avg. words} \\ \midrule
        VT-SSum  & 33.3 & 293.1 & 4208.1 \\

    \bottomrule
    \end{tabular}
    \caption{Data statistics of VT-SSum in video-level}
    \label{tab:1}
\end{table}

\begin{table}[t]
    \centering
    \begin{tabular}{cccc}
    \toprule
        \bf{Dataset} &  \bf{Samples} & \bf{Avg. sents} & \bf{Avg. words} \\ \midrule
        VT-SSum & 125,004  & 12.7 & 185.9 \\
        CNN/DM & 311,971  & 38.6 & 690.9 \\
        
    \bottomrule
    \end{tabular}
    \caption{Data statistics of VT-SSum in page-level compared with CNN/DM}
    \label{tab:2}
\end{table}

\section{Experiment}

\begin{table*}
\centering
\setlength{\tabcolsep}{8mm}
\begin{tabular}{cc}
\hline \textbf{Models} & \textbf{Accuracy} \\ \hline
$\textrm{LSTM}$ \cite{koshorek-etal-2018-text} & 90.33\\
$\textrm{UniLMv2}_{\rm BASE}$ \cite{pmlr-v119-bao20a} & 92.14\\
$\textrm{UniLMv2}_{\rm LARGE}$ \cite{pmlr-v119-bao20a} & 93.00\\
\hline
\end{tabular}
\caption{\label{tab:5} Evaluation result of the segmentation on the test data of VT-SSum }
\end{table*}

\begin{table*}[ht]
\centering
\begin{tabular}{ccccccc}
\toprule \multirow{2}{1.5cm}{\bf Models} & \multicolumn{3}{c}{\bf Top-3}  & \multicolumn{3}{c}{\bf Top-5}  \\ \cline{2-7} 
 & \multicolumn{1}{c}{\bf Precision} & \multicolumn{1}{c}{\bf Recall} & \multicolumn{1}{c}{\bf F1} & \multicolumn{1}{c}{\bf Precision} & \multicolumn{1}{c}{\bf Recall} & \multicolumn{1}{c}{\bf F1}   \\ \midrule
CNN/DM  & 31.49 & 58.41 & 40.92 & 27.10 & 74.46 & 39.74    \\ 
VT-SSum  & 37.86 & 69.04 & 48.90 & 29.79 & 80.79 & 43.53   \\ 
CNN/DM $\to$ VT-SSum &  38.10 &  69.48 &  49.22 &  29.88 &  81.02 &  43.66   \\ 
\bottomrule
\end{tabular}
\caption{Evaluation results on the test data of VT-SSum with models fine-tuned on different datasets} 
\label{tab:3}
\end{table*}

\begin{table*}[ht]
\centering
\begin{tabular}{ccccccc}
\toprule \multirow{2}{1.5cm}{\bf Models} & \multicolumn{3}{c}{\bf Top-3}  & \multicolumn{3}{c}{\bf Top-5}  \\ \cline{2-7} 
 & \multicolumn{1}{c}{\bf Precision} & \multicolumn{1}{c}{\bf Recall} & \multicolumn{1}{c}{\bf F1} & \multicolumn{1}{c}{\bf Precision} & \multicolumn{1}{c}{\bf Recall} & \multicolumn{1}{c}{\bf F1}   \\ \midrule
CNN/DM  & 45.30 & 59.03 & 51.26 & 39.03 & 72.61 & 50.77    \\ 
VT-SSum  & 51.80 & 67.96 & 58.79 & 42.72 & 79.26 & 55.51   \\ 
CNN/DM $\to$ VT-SSum &  52.66 &  68.72 &  59.62 &  42.99 &  79.78 &  55.87   \\ 
\bottomrule
\end{tabular}
\caption{Evaluation results on the test data of AMI with models fine-tuned on different datasets} 
\label{tab:4}
\end{table*}

\subsection{Settings}
We split the VT-SSum into 7,692/962/962 videos for the train, validation and test sets, which includes 99,504/12,569/12,931 training samples in page-level respectively. To verify the effectiveness of VT-SSum on spoken text, we fine-tune the PreSumm model using different datasets: CNN/DM only, VT-SSum only, and 2-stage fine-tuning with CNN/DM first and then VT-SSum. All of these models are evaluated on the test set of VT-SSum. In this way, the results reflect the performance gap between in-domain and out-of-domain training data. Furthermore, we also create a test set from the AMI corpus~\cite{10.1007/11677482_3} for meeting summarization with the input length less than 512, which totally includes 1,337 evaluation samples in spoken language and will be released for future research. 

For the transcript segmentation, we use a Transformer-based pre-trained model UniLMv2~\cite{pmlr-v119-bao20a} to predict \{S,O\} tags for each sentence, where ``S'' indicates that the current sentence is the start of a new segment and ``O'' stands for sentences within the current segment. We use the sliding windows to ensure that all sentences in the transcript can be covered due to the length limitation of the model's input, where the maximum size of a window is 512 and the stride is 384. The overlapping between the windows could make the boundary sentences not lose their contextual information, then we take the result for each sentence with the maximal context. Both base and large models are trained for 30,000 steps with the batch size 32. The learning rate is 5e-5 with a linear warm-up over the first 3,000 steps. All models are trained using 4 NVIDIA V100 16GB GPUs.

During the fine-tuning and testing of summarization, we only keep the first 512 tokens for the VT-SSum and AMI test set for computational efficiency. For the evaluation, we use the F1 score of top-3 and top-5 results as the metric for the extractive summary. We use the ``bert-base-uncased" version of PreSumm to initialize the parameters. All models are trained for 50,000 steps with gradient accumulation every two steps and evaluated on the validation set every 1,000 steps.  We use Adam for optimization with $\beta_{1}$=0.9 and $\beta_{2}$=0.999. The learning rate is 2e-3, with linear warm-up over the first 10,000 steps.



\subsection{Results on Transcript Segmentation}

Table~\ref{tab:5} gives the accuracy of different models on the transcript segmentation dataset. For the baseline, we use a LSTM-based model where the implementation follows the work in \cite{koshorek-etal-2018-text}. It is observed that the LSTM-based model, as a simple but efficient baseline, performs well on the transcript segmentation dataset. Meanwhile, we also use the pre-trained UniLMv2~\cite{pmlr-v119-bao20a} which has more parameters and better performance . It shows that the base and large models of UniLMv2 outperform the LSTM model by 1.81 points and 2.67 points on the accuracy score respectively. These results show the effectiveness of the transcript segmentation models on VT-SSum.


\subsection{Results on Transcript Summarization}

We compare the PreSumm models fine-tuned with CNN/DM and VT-SSum on two different datasets, which are the test set of VT-SSum and a dataset from the AMI corpus. In Table~\ref{tab:3}, it is observed that the model fine-tuned with VT-SSum significantly outperforms the model with CNN/DM with 8 points on Top-3 F1 and 4 points on Top-5 F1, given that the number of training samples are even less than half of the CNN/DM samples. For the 2-stage fine-tuning, the model can further improve based on the model fine-tuned with CNN/DM. This confirms the domain discrepancy between the written language and spoken language on the text summarization task. Furthermore, we also demonstrate the evaluation results on a dataset from the AMI corpus for meeting summarization in Table~\ref{tab:4}. We can see that the VT-SSum model still substantially outperforms the CNN/DM model by a large margin, which improves by 7 points on Top-3 F1 and 5 points on Top-5 F1. We observe that the model trained with CNN/DM often gives lower scores to the sentences with modal particles which are prevalent across the spoken language dataset. This also verifies that the VT-SSum dataset indeed helps the spoken text summarization model. 

\section{Conclusion}

In this paper, we present VT-SSum, a benchmark dataset that contains 125K training pairs for video transcript segmentation and summarization. We adopt a weak supervision method to create the data using videos and slides from VideoLectures.NET. Experiment results show that the model trained with VT-SSum significantly outperforms the the existing models using news datasets on the spoken text summarization task. Besides, for the text segmentation task, the experiment results show the high-quality of the transcript segmentation dataset in VT-SSum. This makes the text summarization models perform much better on the spoken language tasks. 

For future research, as video/meeting summarization has become a new research trend, we will further investigate the joint training of transcript segmentation and summarization to address the lengthy input from videos/meetings. Furthermore, we will investigate how to leverage the abstractive summarization in these scenarios. 

\bibliographystyle{ACM-Reference-Format}
\bibliography{sample-base,anthology}










\end{document}